\pdfoutput=1
\documentclass[11pt]{article}

\usepackage{acl}
\usepackage{amsmath}

\usepackage{times}
\usepackage{latexsym}
\usepackage{multirow}
\usepackage{graphicx}
\usepackage{subfigure}
\usepackage{amsmath}
\usepackage{algorithm}
\usepackage{algpseudocode}
\usepackage{booktabs}
\usepackage{graphicx}

\usepackage[T1]{fontenc}
\usepackage[utf8]{inputenc}

\usepackage{microtype}

\title{Exploring Speaker-Related Information in Spoken Language Understanding for Better Speaker Diarization}

\author{Luyao Cheng, Siqi Zheng, Zhang Qinglin, Hui Wang, Yafeng Chen, Qian Chen \\
Speech Lab of DAMO Academy, Alibaba Group \\
\texttt{\{shuli.cly,zsq174630,qinglin.zql,tanqing.cq\}@alibaba-inc.com}
}

\begin{document}
  \maketitle
\begin{abstract}
Speaker diarization(SD) is a classic task in speech processing and is crucial in multi-party scenarios such as meetings and conversations. 
Current mainstream speaker diarization approaches consider acoustic information only, which result in performance degradation when encountering adverse acoustic conditions. In this paper, we propose methods to extract speaker-related information from semantic content in multi-party meetings, which, as we will show, can further benefit speaker diarization. 
We introduce two sub-tasks, \textbf{Dialogue Detection} and \textbf{Speaker-Turn Detection}, in which we effectively extract speaker information from conversational semantics. 
We also propose a simple yet effective algorithm to jointly model acoustic and semantic information and obtain speaker-identified texts.
Experiments on both AISHELL-4 and AliMeeting datasets show that our method achieves consistent improvements over acoustic-only speaker diarization systems.

% This document is a supplement to the general instructions for *ACL authors. It contains instructions for using the \LaTeX{} style files for ACL conferences. 
% The document itself conforms to its own specifications, and is therefore an example of what your manuscript should look like.
% These instructions should be used both for papers submitted for review and for final versions of accepted papers.
\end{abstract}

\section{Introduction}\label{sec:intro}

Speaker diarization(SD) is the task of answering the question ``who speaks when" by partitioning audio into segments with speaker identities. In most application settings, the results of speaker diarization are perceived by readers through the assignment of speaker labels to the corresponding words or sentences transcribed from an Automatic Speech Recognition(ASR) system.

% pipeline description
Despite the rich profusion of transcribed texts, mainstream speaker diarization systems consider only acoustic information \cite{Park2021ARO, Horiguchi2020EndtoEndSD, Park2020AutoTuningSC, Fujita2019EndtoEndNS, DBLP:conf/icassp/ZhengHWSFY21, DBLP:journals/corr/abs-2203-09767}. Traditional SD systems usually consist of the following components: 
(1) Voice activity detection (VAD) to filter out non-speech frames. (2) Extraction of speaker embeddings from the short audio segments, using popular models such as i-vector\cite{Dehak2011FrontEndFA}, d-vector\cite{Zhang2017EndtoEndTS} and x-vector\cite{Snyder2018XVectorsRD}. (3) 
Clustering embeddings into several classes using algorithms such as agglomerative hierarchical clustering (AHC) \cite{Day1984EfficientAF}, spectral clustering(SC) \cite{Wang2017SpeakerDW}, and HDBSCAN \cite{DBLP:conf/interspeech/ZhengSC22}. Various speaker embedding model and clustering methods have been explored and proposed in  \cite{DBLP:conf/icassp/YuZSLL21, Dawalatabad2021ECAPATDNNEF, He2021TargetSpeakerVA,DBLP:conf/icassp/ZhengS22,DBLP:journals/corr/abs-2211-10243}.
% EEND may be too much?
% End-to-end(E2E) joint optimization based models have also been applied to speaker diarization task.
% EEND\cite{Fujita2019EndtoEndNS}, a typical example E2E method, forms speaker diarization as a multi-label classification problem using permutation invariant training\cite{DBLP:journals/corr/abs-1909-05952}. EEND can only solve the scenario where the number of speakers is limited, and the subsequent EEND-EDA\cite{horiguchi2020end} and SC-EEND\cite{fujita2020neural} models have explored this problem. Nevertheless, the E2E model still needs to be improved in the speaker diarization task where the number of speakers is unknown.

% exist text for speaker diarization methods
Utilizing only acoustic information has significant limitations. For example, the performance of SD system suffers from obvious degradation in adverse acoustic conditions such as noise, reverberation, and far-field recordings. In addition, we often encounter speakers with similar voice characteristics, which pose serious challenge to clustering them into expected classes. Given the abundance of transcribed texts present in meetings and conversations, it is of sufficient interest to explore the possibilities of utilizing semantic information to go beyond the limits of acoustic-only speaker diarization. 

Some previous works tried to use semantic information to classify roles in two-speaker conversations, such as doctor-patient conversation and pilot-air traffic controller dialogue\cite{ZuluagaGomez2021BERTrafficBJ, Flemotomos2022MultimodalCW}. However, these methods are only suitable for specific two-speaker scenarios where the roles are clearly-defined, such as medical diagnosis, job interviews, and air traffic communications. In this work we focus on open multi-party meeting scenarios where the number of speakers is unknown and the relations between speakers are unspecified. 

\begin{figure*}[ht!]
    \centering
    \includegraphics[width=1.0 \textwidth]{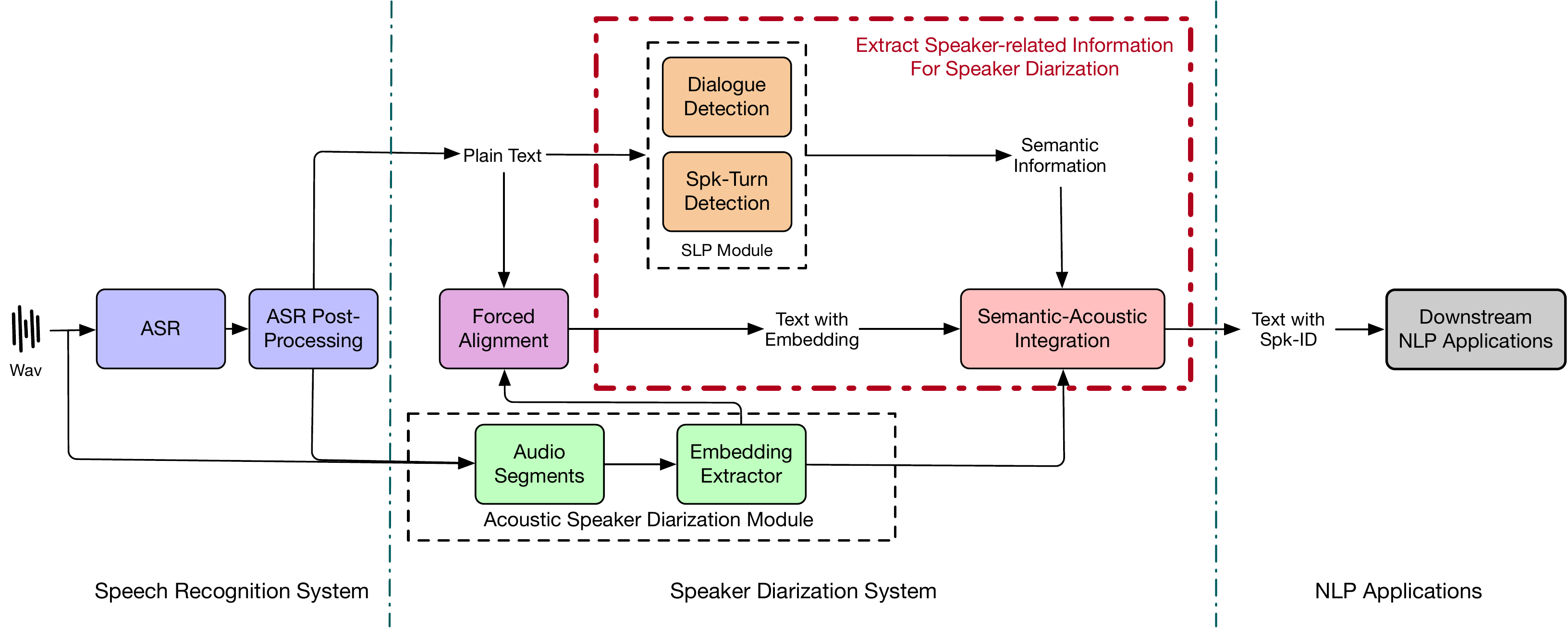}
		%\centerline{\includegraphics[width=10.0cm]{figures/topo}}  %,trim=0 40 0 0
    %\setlength{\abovecaptionskip}{-1.5cm}
    %\vspace{-0.2cm}
    \caption{We propose a multi-modal speaker diarization system that utilizes the SLP module to extract speaker-related information from transcribed text. The multi-modal fusion and semantic backend modules combine both acoustic and semantic information to improve the accuracy of speaker diarization. The system's output includes text segments with corresponding speaker identification.}
    \label{fig:sd-system}
\end{figure*}

Speaker identity information has been proven to be beneficial to many downstream NLP tasks\cite{chi2017speaker, zhang2018addressee, Chi2017SpeakerRC}. However, these works only consider speaker identities as given ground truth \cite{carletta2006ami, Janin2003TheIM, Zhong2021QMSumAN}, which is impractical in real world settings. 
Therefore, it is crucial to make valid inference of speaker identities on the transcribed conversations using a well-performed speaker diarization system. 

% contribution
The main contributions of this paper include: 

(1) We propose two semantic tasks to extract speaker-related information from automatically transcribed texts, namely Dialogue Detection and Speaker-Turn Detection.

(2) We design a simple yet effective integration method to effectively combine semantic and acoustic information for more robust speaker diarization.

\section{Proposed Methods}

\subsection{A Novel Multi-modal Framework}
Figure \ref{fig:sd-system} illustrates the proposed semantic-acoustic speaker diarization system, along with its relation with upstream ASR components and downstream NLP applications. We introduce a Spoken Language Processin(SLP) module involving two sub-tasks to extract speaker-related information from transcribed texts. The acoustic-based speaker diarization system is used to process original audio, perform segmentation and estimate speaker embeddings for each segments. To associate speaker embeddings with corresponding text phrases, a forced alignment component was introduced to our system. Finally, we propose an integration method to collectively process outputs from SLP module, acoustic SD module, and forced alignment module. 

\subsection{Learning Speaker Information From Texts}

To extract semantic speaker-related information, 
we define two sub-tasks: \textbf{dialogue detection} and \textbf{speaker-turn detection}. 

%Let $X = \{x^{(1)}, x^{(2)}, ..., x^{(N)}\}$ be a sequence of transcribed sentences, then we represent the transcribed texts from the entire meeting 
 %We use overlapping-sliding-window method to sentence sequence $\{X_1, X_2, ..., X_K\}$.

Dialogue detection takes a sequence of sentences as input and determines whether this is transcribed from a multi-speaker dialogue or a single-speaker speech. Dialogue-detection can be defined as a binary classification problem.
%\begin{equation}
%    \hat{Y}_{dd}^k = \mathop{argmax}\text{P}(Y_{dd}|X_s^{k}), C \in \{0, 1\}
%\end{equation}
% For dialogue detection task, we need to control the length of the text we choose because in a multi-person dialogue scenario, the longer the text, the higher the possibility of containing multiple speakers, so the text that is too long is meaningless for the following experiments.

Speaker turn detection tries to predict, for each given sentence in the sequence, the probability of the occurrence of speaker change. Speaker turn detection can be defined as a sequence labeling problem, where the goal is to determine whether the given position represents a point of change in speaker role from a semantic perspective.

Both dialogue detection and speaker turn detection models are fine-tuned from a pre-trained BERT language model. Design of training samples and details of experiments are discussed in next section.

%\begin{equation}
%    \hat{Y}_{std} = \mathop{argmax}\text{P}(Y_{std}|X_s^{k}), Y \in \{0, 1\}^{n_{seg}}
%\end{equation}

% To reduce the sparsity of speaker turn labels, we use punctuation to segment the sentence $X_s^{k}$ into $n_{seg}$ segments $S^k = \{s_1^k, s_2^k, ..., s_{n_{seg}}^k\}$. The task is to predict whether segmentation boundary $Y_{scp}^k = \{y_1^k, y_2^k, ..., y_{n_{seg}}^k\}, y_i^k \in \{0, 1\}$ is a speaker turn, and $y_i^k = 1$ means between $s_i^k$ and $s_{i+1}^k$ is a speaker turn.

% The Punctuation information can help effectively, which means only judging whether it is a speaker-turn at the punctuation position.

\subsection{Integrating Semantic-Acoustic Information}

In this section we describe how speaker-related information extracted from semantic content can assist us in improving upon acoustic-only SD system. A traditional SD system typically involves an audio segmentation module and an embedding clustering module. Poor segmentation and incorrect clustering are the most common problems in speaker diarization. Semantic information from dialogue detection helps improve clustering accuracy and speaker turn detection helps to find more precise place in text where a change of speaker occurs.

%To integrate multi-modal results, we first utilize a forced alignment module to associate each speaker embedding with the corresponding text phrase. 

Note that dialogue detection and speaker turn detection tasks can be solved either by acoustic-only approach or semantic-only approach. Semantic-only approach is described above. Acoustic-only results can be derived directly from acoustic-based speaker clustering. The speaker clustering algorithm assigns a cluster label to each speaker-segment. Acoustic results for dialogue detection can be obtained simply by checking whether the number of different speaker labels is larger than 1. Results for speaker turn detection can be obtained by analyzing the transition patterns of speaker-segment labels or predicting change points using an acoustic-based neural networks such as Target-Speaker VAD\cite{He2021TargetSpeakerVA}. 

\textbf{Semantic-Acoustic Dialogue Detection.} Let $z^{(s)}$ denote the result of binary classification output of semantic dialogue detection and $z^{(a)}$ be the counterpart of acoustic dialogue detection. We also define $D_p$ to be the distance of the largest speaker cluster present in the dialogue to its furthest cluster, and $D_q$ to be the standard deviation of the cosine distances among all speaker embeddings present in the selected speech. $D_p$ measures how spread out different clusters are and $D_q$ measures how tight embeddings in one cluster are grouped together. Then the fusion score $\hat{s}$ is estimated by:

\begin{equation}
\hat{s}= z^{(a)}z^{(s)} +z^{(a)}(p_s + \alpha_1 D_p) +z^{(s)}(p_s + \alpha_2 D_q),
\end{equation}

where $\alpha_1$ and $\alpha_2$ are learnable and $p_s$ is logit output from semantic dialogue detection.

For some threshold $\theta$, the binary output of semantic-acoustic dialogue detection is represented by the indicator function:

\begin{equation}
    \begin{aligned}
        \hat{z}^{\text{fusion}}_{dd}=\mathbf{1}_{\hat{s}>\theta}
    \end{aligned}
\end{equation}

Once semantic-acoustic dialogue detection obtain results for all sentence sequences that cover the entire transcribed meeting, we re-adjust the acoustic-based clustering results. By doing this we are able to incorporate semantic information to improve speaker clustering. More details can be found in Appendix \ref{sec:smb}.

% Among them, $p_a^{\text{split}}$ and $p_a^{\text{merge}}$ are calculated based on the distribution of the embedding in the sub-sentence, where $p_a^{\text{split}}$ is the maximum value of the maximum cluster distance calculated in the case of z=1, and $p_a^{\text{merge}}$ is the standard deviation of the cosine distance when z=0.

% $p_a^{\text{split}}$ and $p_a^{\text{merge}}$ are calculated based on the distribution of the embeddings in the sub-sentence. 

\textbf{Semantic-Acoustic Speaker Turn Detection.}  Semantic-only speaker turn detection outputs a sequence of probability of the occurrence of speaker change. Let $p_n$ be the probability at position $n$, and $q_n$ represents the speaker change probability from an acoustic-only model near position $n$. $q_n$ is obtained by taking the maximum probability of the closest 200 frames estimated by the Target-Speaker VAD model. Then the integrated speaker-change probability is given by

\begin{equation}
    \Tilde{p}_{n} = \beta_1 p_n + \beta_2 q_n
\end{equation}

for some learnable hyperparameters $\beta_1$ and $\beta_2$.

% The text $X_{s}$ is cut to $D = \{d_s^1, d_s^2, ..., d_s^{N_D} \}$ and each $d_s^i$ is a short sentence cutted by punctuation and VAD.
% We modified the speaker diarization pipeline, as shown in the Fig., the speaker diarization task will be executed after the text recognition task in order to obtain the semantic information contained in the speech. Using the forced-alignment model, we can get the time of each short sentence, and then obtain the speaker-embedding $E_i = \{e_1^i, e_2^i, ..., e_{N_i}^i\}$ contained in each short sentence $d_s^i$.

%\subsection{Semantic Backend}
% We design a post-processing algorithm that utilizes information obtained from the two subtasks to adjust the speaker diarization result.

%\textbf{Split-Merge Process}
% TODO
%Given a set of short sentences, $W = \{w_1, w_2, ..., w_{W_N}\}$, and a set of constraints, $P = \{p_1, p_2, ..., p_{W_N-1}\}$, indicating whether each adjacent pair of sentences, $w_i$ and $w_{i+1}$, come from the same speaker. We use a split-merge method to adjust the speaker identification for each short sentence.
%In the split process, we increase the number of speakers as necessary to ensure the speaker identification is consistent with the speaker-related constraints. 
%In the merge process, we consider the similarity of the embeddings of each speaker and the impact on the constraint set accuracy before determining whether to merge two speaker IDs.
%More details can be found in Appendix \ref{sec:smb}

\textbf{Boundary and Outlier Correction.}
We use semantic information to correct boundary errors caused by errors and mismatches from the forced-alignment and ASR models. We also use semantic information to correct outliers in embedding extraction. To improve system robustness, we exclude audio segments that are too short from clustering. Outliers and left-out embeddings are assigned to the closest cluster.

\section{Experiments and Results}
% In this section, we compare the performance of our method with classic recipes. 

\subsection{Datasets}
% Datasets
We conduct experiments on AISHELL-4\cite{fu2021aishell} and AliMeeting\cite{9746465} datasets. Both focus on multi-party meeting scenario, where all speech content are manually annotated. Table \ref{tab:data-situation} listed detailed information about the datasets. We perform experiments using both ground truth (GT) text and text transcribed from ASR system.

% \usepackage{booktabs}
% \usepackage{graphicx}
\begin{table}[]
\footnotesize
\resizebox{\columnwidth}{!}{%
\begin{tabular}{@{}ccc@{}}
\toprule
 & \begin{tabular}[c]{@{}c@{}}AISHELL-4\\ Train/Eval\end{tabular} & \begin{tabular}[c]{@{}c@{}}Alimeeting\\ Train/Eval\end{tabular} \\ \midrule
Session             & 191/20          & 209/20            \\
\#Avg. Duration(s)  & 1939.03/2245.94 & 1915.52/1924.7    \\
\#Avg. Speakers     & 4.8/5.8         & 3.27/3.0          \\
\#Avg. Speaker-Turn & 343.95/220.8    & 649.34/552.75     \\
Avg. Text Len.      & 8904.8/9990.9   & 13249.55/12067.05 \\ \bottomrule
\end{tabular}%
}
\caption{Details of AISHELL-4 and AliMeeting data.}% The labels for dialogue detection and speaker turn detection are built from speaker labels in the annotation. We construct the semantic training data by combining the AISHELL-4 and Alimeeting training sets.}
\label{tab:data-situation}
\end{table}

\begin{table*}[ht]
    \centering
    \footnotesize
    % \resizebox{\columnwidth}{!}{%
    \begin{tabular}{@{}ccccccccc@{}}
    \toprule
    \multirow{2}{*}{TaskName}                                                         & \multirow{2}{*}{Text}                    & \multirow{2}{*}{Methods} & \multicolumn{3}{c}{AISHELL-4}        & \multicolumn{3}{c}{Alimeeting}                                     \\ \cmidrule(l){4-9} 
                                                                                      &                                          &                          & Precision & Recall & F1              & Precision            & Recall               & F1                   \\ \midrule
    \multirow{6}{*}{\begin{tabular}[c]{@{}c@{}}Dialogue\\ Detection\end{tabular}}     & \multirow{3}{*}{GT}                      & Acoustic-Only                 & 74.402    & 84.995 & 79.346          & 93.012               & 92.259               & 92.634               \\
                                                                                      &                                          & Semantic-Only                 & 74.649    & 96.976 & 84.360           & 94.669               & 98.009               & 96.310                \\
                                                                                      &                                          & Multi-Modal                   & 86.308    & 93.402 & \textbf{89.715} & 96.450                & 97.600                 & \textbf{97.020}       \\ \cmidrule(l){2-9} 
                                                                                      & \multirow{3}{*}{ASR}                     & Acoustic-Only                 & 80.405    & 96.936 & 87.900            & 96.482               & 98.428               & 97.445               \\
                                                                                      &                                          & Semantic-Only                 & 55.731    & 84.414 & 67.138          & 93.649               & 88.688               & 91.101               \\
                                                                                      &                                          & Multi-Modal                   & 82.461    & 95.826 & \textbf{88.642} & 96.641               & 98.320                & \textbf{97.476}      \\ \midrule
    \multirow{6}{*}{\begin{tabular}[c]{@{}c@{}}Speaker-Turn\\ Detection\end{tabular}} & \multicolumn{1}{l}{\multirow{3}{*}{GT}}  & Acoustic-Only                 & 53.962    & 51.272 & 52.583          & 54.329               & 52.997               & 53.655               \\
                                                                                      & \multicolumn{1}{l}{}                     & Semantic-Only                 & 69.569    & 89.514 & 78.291          & 76.696              & 93.141              & \textbf{84.123}     \\
                                                                                      & \multicolumn{1}{l}{}                     & Multi-Modal                   & 81.652    & 77.240  & \textbf{79.385} & 76.861               & 92.849               & 84.102               \\ \cmidrule(l){2-9} 
                                                                                      & \multicolumn{1}{l}{\multirow{3}{*}{ASR}} & Acoustic-Only                 & 61.657    & 79.162 & 69.322          & 66.105              & 67.696              & 66.891              \\
                                                                                      & \multicolumn{1}{l}{}                     & Semantic-Only                 & 42.299    & 63.386 & 50.738          & 61.046              & 50.100              & 55.034              \\
                                                                                      & \multicolumn{1}{l}{}                     & Multi-Modal                   & 68.132    & 73.878 & \textbf{70.889} & 67.593               & 66.960                & \textbf{67.276}      \\ \bottomrule
    \end{tabular}%
    % }
    \caption{The results of two sub-tasks on AISHELL-4 and Alimeeting test set.}
    \label{tab:semantic-result-label}
    \end{table*}

\begin{table*}[ht!]
\centering
\footnotesize
% \resizebox{\columnwidth}{!}{%
\begin{tabular}{@{}cccccccc@{}}
\toprule
\multirow{2}{*}{Text Type} & \multirow{2}{*}{Methods} & \multicolumn{3}{c}{AISHELL-4} & \multicolumn{3}{c}{Alimeeting} \\ \cmidrule(l){3-8} 
                           &                          & cp-wer & cp-wer-all & speaker-wer & cp-wer  & cp-wer-all & speaker-wer \\ \midrule
\multirow{3}{*}{GT Text}   & Baseline - only acoustic info.                                & 17.309  & 19.099     & \textbf{5.974}    & 41.669  & 52.617   & 18.888   \\
                           &  Semantic-acoustic info. - A            & 15.540  & 18.798      & 6.558    & 36.360   & 45.772     & 14.700   \\
                           & Semantic-acoustic info. - B              & \textbf{15.225} & \textbf{18.364}  & 6.281  & \textbf{36.145}  & \textbf{45.462}   & \textbf{14.500}    \\ \midrule
                           % & Backend with Fusion Info.               &        &            &         &         &            &         \\ \midrule
\multirow{3}{*}{ASR Text}  & Baseline - only acoustic info.                               & 33.905   & 35.590    & 3.647    & 45.678   & 49.778     & 8.404     \\
                           & Semantic-acoustic info. - A             & 33.355  & 34.318    & 2.650    & 38.467   & 40.182      & 2.413   \\
                           & Semantic-acoustic info. - B               & \textbf{33.290}   & \textbf{34.210}     & \textbf{2.575}   & \textbf{38.459} & \textbf{40.154}  & \textbf{2.389}    \\ \bottomrule
                           % & Backend with Fusion Info.               &        &            &         &         &            &         \\ \bottomrule
\end{tabular}%
% }
\caption{The speaker diarization results of different systems on the AISHELL-4 and Alimeeting test set}
\label{tab:cpwer-results}
\end{table*}

% \subsection{Acoustic Modules}
% The acoustic recipe can be divided into following components: 
% (1) \textit{Speaker Diarization module}. In our experiments, The speaker embedding extractor is based on \citealp{Desplanques2020ECAPATDNNEC}, and the speaker clustering alogrithm is Spectral Clustering.
% (2) \textit{ASR module}. In our experiments, This module consists of ASR and punctuation prediction models and they are all Mandrain models.
% (3) \textit{Forced-Alignment module}. We apply forced-alignment module is to find the timestamps for each character from the text. Note that the forced alignment could cause some deletion errors, so we apply Needleman-Wunsch algorithm\cite{NEEDLEMAN1970443} to match the text. \textbf{All these components are fixed in all our experiments.}
% Semantic information is used in the process of segmenting speech to extract speaker embedding and obtaining text with speaker id from speaker clustering and forced alignment results.

\subsection{Experimental Setups}
In our experiments, the acoustic modules, including ASR, ASR Post-Processing, Embedding Extractor, and Forced Alignment models, are fixed and used consistently throughout all our experiments. In details, the ASR system we introduced was based on UniASR\cite{Gao2020UniversalAU}. The ASR Post-Processing contained Punctuation-Prediction\cite{Chen2020ControllableTT} and Text-Smoothing which are common used in meeting scenrio. The Forced Alignment module we introduced was based on \cite{McAuliffe2017MontrealFA}. For acoustic speaker diarization system, we employed a speaker embedding extractor based on ECAPA-TDNN\cite{Desplanques2020ECAPATDNNEC}, while for speaker clustering, we utilized Spectral Clustering algorithm with p-percentile\cite{Wang2017SpeakerDW}.

We fine-tune the semantic models for dialogue detection and speaker turn detection tasks based on the pre-trained BERT language model\footnote{Based on bert-base-chinese from HuggingFace} using the text from AISHELL-4 and AliMeeting training sets. Training samples are sequences of sentences generated by a sliding-window method with a window length of 64 and a shift of 16 and the label for these two semantic subtasks can be generated by the speaker label from the speech content manually annotated in the datasets.
% We randomly divide the concatenated text into train and test sets using a sliding-window method. To maintain semantic integrity, we only segment at punctuation positions with a window length of 64 and a shift of 16. If the segmentation point is not at a punctuation position, it will be filled to the next punctuation position.

For dialogue detection task, we fine-tune for 3 epochs on train dataset with a learning rate of 5e-6 and a batch size of 64. For speaker-turn detection task, we also fine-tune for 3 epochs on train dataset with a learning rate of 1e-6 and a batch size of 64.

%For dialogue detection task, we fine-tune 3 epochs on the mixture train dataset with learning rate 5e-6 and batch size is 64. For speaker-turn detection tasks, %to reduce the sparsity of speaker turn labels, we only consider speaker-turn points after punctuation. % There is no problem in the GT text, but it will introduce some mistakes in the ASR text as there are text errors and the punctuation results are also not completely correct. 
% we also fine-tune 3 epochs on the mixture train dataset with learning rate 1e-6 and batch size is 64.

\subsection{Results and Discussions}
We compare our proposed methods with the classic speaker diarization system mentioned in Section \ref{sec:intro}.

% Table \ref{tab:semantic-result-label} shows results of Dialogue Detection and Speaker-Turn Detection tasks based the acoustic methods and the semantic method. The semantic model based on the pre-trained model can well complement the pure acoustic model, so as to better model the information of local speaker turn to obtain the text with speaker label.

Table \ref{tab:semantic-result-label} shows the results of dialogue detection and speaker-turn detection tasks from acoustic-only, semantic-only, and multi-modal models. We not only compare results using ASR-transcribed texts, but also conduct experiments using ground truth texts as inputs, in order to see the optimal improvements introduced by semantic information. The multi-modal model surpasses single-modal results on both GT and ASR text. The experiments demonstrate that semantic model can effectively supplement acoustic-only model, resulting in more precise speaker representation. It is expected that the introduction of semantic information on ASR text does not result in a significant improvement due to the a lot of errors present in the text. However, our multi-modal approach shows consistent improvement in both GT and ASR-based results, indicating the robustness of our method.
% The results also show that the performance of the SLP model on the ASR text is significantly worse than that of the GT text. 

% \subsection{Speaker Diarization Results}
% add figure to show

We use the \textbf{cp-WER} metric to measure the speaker diarization task. We introduce a new metric \textbf{speaker-WER} that aims to measure the word error rate caused \textbf{solely} by speaker errors. More details about the metrics can be found in Appendix \ref{sec:appendix-metrics}. Table \ref{tab:cpwer-results} shows the final results of our speaker diarization system. Compared to the baseline, which only uses acoustic information, our system incorporating semantic information shows a significant improvement. The results for ``Semantic-acoustic info. - A" indicate that only semantic information is used for sub-tasks Dialogue Detection and Speaker Turn Detection, while ``Semantic-acoustic info. - B" indicates that both semantic and acoustic information are used in the two sub-tasks.

% \section{Discussion}
% \subsection{Purity of Speaker Embedding}
% Analyzing from the perspective of speaker change point, these semantic punctuation marks are taken as candidate speaker change point. The advantage of this method is that it reduces the number of speaker embedding from speaker overlap or speaker confusion, and improves the purity of speaker embedding for the following speaker clustering process.

\section{Conclusions}
We propose a novel multi-modal speaker diarization system that utilizes two spoken language processing tasks, dialogue detection and speaker-turn detection, to extract speaker-related information from text. These information are then combined with acoustic information to improve the overall performance of speaker diarization. Our experiments demonstrate that incorporating semantic information can effectively address the limitations of single-modal speech.

\section{Limitations}
The performance of SLP tasks rely heavily on the accuracy of ASR system. Poorly-transcribed texts can lead to degradation of our multi-modal method. Since we cannot easily obtain accurate speaker-turn information from the ASR text, the training set for SLP tasks based on ASR text is also not easy to obtain. In future work, we will try more methods, like Data Arguments, to get better results on ASR text. 

Overlapping speech is another challenge for the task, as a monaural ASR system can no longer capture all spoken words from all speakers. In future work, we plan to explore methods such as speech separation or multi-party ASR to handle overlapping speech.

\bibliography{anthology,custom}
\bibliographystyle{acl_natbib}

\appendix

\section{Metrics} \label{sec:appendix-metrics}
% For Appendix maybe?
% describe cpWER
The performance metrics for speaker diarization is concatenated minimum-permutation Word error rate(cpWER). The cpWER is computed as follows:
\begin{enumerate}
    \item Concatenate each speaker's utterances both from reference and hypothesis results.
    \item Compute the WER between the reference and all possible speaker permutation of hypothesis.
    \item Choose the lowest WER among the results from all the speaker permutations as the final cpWER.
\end{enumerate}
Specifically, the Word Error Rate(WER) is calculated by: 
\begin{equation}
    E_{wer} = \frac{N_{Ins} + N_{Subs} + N_{Del}}{N_{Total}} \times 100\%
\end{equation}

The cpWER is affected both by the speaker diarization system and speech recognition system. 

Note that the number of speakers in the system result and the reference result are not equal. If we ignore the cpWER measured by this part of the text, we will record it as $E_{cp-matched}$. If we think that this part of the text should be considered all wrong, we will record the cpWER this time for $E_{cp-all}$.

We know that WER calculates the minimum edit distance from the system result to the reference result. In the calculation process, the cost of changing one sequence into another sequence by using three operations of insertion, deletion and replacement is counted, while the calculation of cpWER additionally introduces the operation of modifying the speaker ID of a word to make two speaker-labeled texts become consistent. 

Since the speaker diarization system cannot modify the text results of speech recognition, we calculate the speaker-WER by removing the errors caused by ASR results from cpWER. Compared to the three string operations in WER, we additionally define an operation to convert one text result with speaker ID to another by modifying the speaker ID of a word. Similar to the WER algorithm, we use dynamic programming to count the number of operations for changing the speaker ID. The speaker-WER results is calculated by:
\begin{equation}
    E_{speakerWER} = \frac{N_{Spk-Cost}}{N_{Total}}
\end{equation}

\section{Pseudocodes to Update Speaker Diarization from Semantic results}\label{sec:smb}
The following psuedocodes show how results from Dialogue Detection and Speaker Turn Detection are utilized to re-adjust and update speaker diarization results.

We create a sequence of speaker change occurrences, $P_{stp} = \{p_1, p_2, ..., p_{N}\}$, where $N$ is the number of speaker changes, by combining the results of the Dialogue Detection and Speaker-Turn  Detection tasks.

We propose a split process, as shown in Algorithm 1, to adjust the speaker IDs of certain segments to align the cluster results with the speaker change results as closely as possible by increasing the number of speakers appropriately.

\begin{algorithm}\label{alg:ssp}
\caption{The Semantic Split Process}
\begin{algorithmic}
\Require $P_{stp}=\{p_1, p_2, ..., p_{N}\}$ The set of the speaker turn point, $p_i \in \{0,1\}$ \\
         $D=\{d_1, d_2, ..., d_N\}$ Each text segments divided by speaker change, \\ 
         $E=\{e_1, e_2, ..., e_N\}$ Set of mean embedding belong to $d_k$, \\
         $\tau_{split}$ The split threshold
\Ensure 
\State $\hat{S} \gets [1]$, $B \gets [e_1]$, $N_{spk} \gets 1$, $i \gets 2$
\While{$i \le N$}
\State $\text{dist} = f(e_i, B) \in R^{\mathop{Size}(B)}$
\If{$p_i = 1$}
    \If{$\text{dist} < \tau_{split}$}
        \State $s_i = \mathop{argmin}_{s_i \in S} \text{dist}$
        \State $\hat{S} \gets \hat{S}\text{.append}(s_i)$
        % \State $N \gets \frac{N}{2}$  \Comment{This is a comment}
    \Else 
        \State $N_{spk} \gets N_{spk} + 1$
        \State $s_i \gets N_{spk}$
        \State $\hat{S} \gets \hat{S}\text{.append}(s_i)$
    \EndIf
\Else
    \State $s_i \gets s_{i-1}$
    \State $\hat{S} \gets \hat{S}\text{.append}(s_i)$
\EndIf
\State $B \gets B \cup e_i$
\EndWhile
\end{algorithmic}
\end{algorithm}

After the split process, a merge process is implemented to eliminate redundant speaker IDs. We consider both acoustic information, such as the similarity distance between speaker embeddings, and semantic information, such as the score differences between the merge of speakers $i$ and $j$, computed from the Dialogue-Detection and Speaker-Turn Detection results of the utterances with the speaker IDs in two adjacent text segments. The pseudocode for the merge process is outlined in Algorithm 2.
\begin{algorithm}
\caption{The Semantic Merge Process}\label{alg:smp}
\begin{algorithmic}
\Require $E$ Set of mean embedding of all the speakers\\
        $S$ Set of speakers after split process \\
        $\tau_{\text{merge}}$ the merge threshold
% \Ensure $y = x^n$
\While{\text{Can Not Merge}}
\State $\text{cost}_{sim} \gets \mathop{cosine\_distance}(E)$
\State $\text{cost}_{dd} \gets \text{score}_{s}^{\text{after}} - \text{score}_{s}^{\text{before}}$
\State $i_{merge}, j_{merge} = \mathop{argmin}_{i,j \in S} (\text{cost}_{sim} + \text{cost}_{s})$
\If{$ \text{cost}_{\text{all}}[i_{merge}, j_{merge}] < \tau_{merge} $}
    \State \text{update} $S \text{ by merge speaker } i_{merge}, j_{merge}$
    \State \text{update} $E \text{ by merge speaker } i_{merge}, j_{merge}$
\EndIf
\EndWhile
\end{algorithmic}
\end{algorithm}

\end{document}